\documentclass[8pt]{article}

\usepackage{amssymb} % \mathbb, etc.
\usepackage{amsfonts}
\usepackage{amsmath} % \equation, \align, etc.
\usepackage{cite}
\usepackage{bm} % bold italic math: \bm
\usepackage{graphicx} % \includegraphics, etc.
\usepackage{cite} % combine citations
\usepackage{authblk} % affiliations
\usepackage{algorithmic}
\usepackage{algorithm}
\usepackage{float}
\usepackage{url}
\usepackage{mathtools}
\usepackage{multirow}
\usepackage{booktabs}
\usepackage{fullpage}
\usepackage{charter} % change font

\graphicspath{{figures/}}

\def\T{{\textsf{\tiny T}}}
\def\defn{\,\triangleq\,} % definitions: equality with triangle
\def\e{\mathrm{e}} % Euler's number
\def\sgn{\mathrm{sgn}} % imaginary number

\def\argmin{\mathop{\mathrm{arg\,min}}} % Argument of a minimization

\def\sbf{\mathbf{s}}
\def\rbf{\mathbf{r}}
\def\ubf{\mathbf{u}}

\def\xbf{\mathbf{x}}

\def\ybf{\mathbf{y}}
\def\cbf{\mathbf{c}}
\def\ebf{\mathbf{e}}
\def\bbf{\mathbf{b}}
\def\vbf{\mathbf{v}}

\def\zbf{\mathbf{z}}
\def\Abf{\mathbf{A}}
\def\Hbf{\mathbf{H}}

\def\Dbf{\mathbf{D}}
\def\Phibf{\mathbf{\Phi}}
\def\Wbf{\mathbf{W}}

\def\Ibf{\mathbf{I}}

\def\xbfhat{\widehat{\mathbf{x}}}

\def\phibm{\bm{\phi}}

\def\thetabm{\bm{\theta}}

\def\Ecal{\mathcal{E}}

\def\Ccal{\mathcal{C}}

\def\Tcal{\mathcal{T}}

\def\Hcal{\mathcal{H}}

\def\Rcal{\mathcal{R}}
\def\Tcal{\mathcal{T}}

\def\R{\mathbb{R}}

\def\Z{\mathbb{Z}}

\def\T{\text{T}}

\usepackage{xspace}
\usepackage{xcolor}

\begin{document}

\title{Learning-based Image Reconstruction via \\ Parallel Proximal Algorithm}

\author{Emrah~Bostan%
\thanks{E.~Bostan (email: bostan@berkeley.edu) 
and L.~Waller (email: waller@berkeley.edu) are with the Department of 
Electrical Engineering \& Computer Sciences, University of California, 
Berkeley, CA 94720, USA. \\ 
E. Bostan's research is supported by the Swiss National Science Foundation
(SNSF) under grant P2ELP2 172278.},
Ulugbek~S.~Kamilov%
\thanks{U.~S.~Kamilov (email: kamilov@wustl.edu) is with Computational Imaging
Group (CIG), Washington University in St.~Louis, St.~Louis, MO 63130, USA.}, and
Laura~Waller$^*$
}

\maketitle %% required

\begin{abstract}
In the past decade, sparsity-driven regularization has led to
advancement of image reconstruction algorithms. Traditionally, such
regularizers rely on analytical models of sparsity (e.g.  total variation
(TV)). However, more recent methods are increasingly centered around
data-driven arguments inspired by deep learning. In this letter, we propose to
generalize TV regularization by replacing the $\ell_1$-penalty with an
alternative prior that is trainable. Specifically, our method learns the prior
via extending the recently proposed fast parallel proximal algorithm (FPPA) to
incorporate data-adaptive proximal operators.  The proposed framework does not
require additional inner iterations for evaluating the proximal mappings of the
corresponding learned prior. Moreover, our formalism ensures that the training
and reconstruction processes share the same algorithmic structure, making the
end-to-end implementation intuitive.  As an example, we demonstrate
our algorithm on the problem of deconvolution in a fluorescence microscope.
\end{abstract}

%%%%%%%%%%%%%%%%%%%%%%%%%%%%%%%%%%%%%%%%%%%%%
%% Sections
%%%%%%%%%%%%%%%%%%%%%%%%%%%%%%%%%%%%%%%%%%%%%

%%%%%%%%%%%%%%%%%%%%%%%%%%%%%%%%%%%%%%%%%%%%%
%% Introduction
%%%%%%%%%%%%%%%%%%%%%%%%%%%%%%%%%%%%%%%%%%%%%

\section{Introduction}
\label{Sec:Intro}

The problem of reconstructing an image from its noisy linear
observations is fundamental in signal processing. 
Formulating the reconstruction as a linear inverse problem
\begin{equation}
\ybf = \Hbf\xbf + \ebf,
\end{equation}
the unknown image $\xbf \in \R^N$ is computed from measurements
$\ybf \in \R^M$. Here, the matrix ${\Hbf \in \R^{M \times N}}$ models the
response of the acquisition device, while $\ebf \in \R^M$ represents the
measurement noise. In practice, the
reconstruction often relies on the regularized least-squares approach:
\begin{align}
\label{Eq:RegularizedOptimization}
\xbfhat = \argmin_{\xbf \in \R^N} \left\{\frac{1}{2}\|\ybf -
\Hbf\xbf\|_{\ell_2}^2 + \tau \Rcal(\xbf)\right\}\hspace{-0.23em},
\end{align}
where $\Rcal$ is a regularization functional that promotes some desired
properties in the solution and  $\tau > 0$
controls the strength of the regularization. 

In most reconstruction schemes, an analytical prior model is used. One of the
most popular regularizers for images is total variation
(TV)~\cite{Rudin.etal1992}, defined as $\Rcal_{\rm TV}(\xbf) \defn
\|\Dbf\xbf\|_{\ell_1}$, where $\Dbf$ is the discrete gradient operator. The TV
functional is a sparsity-promoting prior (via the $\ell_1$-norm) on the image
gradient. Used in compressed sensing~\cite{Donoho2006,Candes.Tao2005}, TV
regularization has been central to inverse problems and successfully applied to
a wide range of imaging applications~\cite{Persson.etal2001,
Bronstein.etal2002, Lustig.etal2007, Kamilov.etal2016}. 

Two commonly used methods for performing TV regularized image reconstructions
are the (fast) iterative shrinkage/thresholding algorithm
((F)ISTA)~\cite{Beck.Teboulle2009a} and alternating direction method of
multipliers (ADMM)~\cite{Afonso.etal2010}.  These algorithms reduce the
complex optimization problem to a sequence of simpler operations applied to the
iterates. Both methods require evaluating the proximal mapping of
the TV regularizer at each iteration~\cite{Moreau1965}. This amounts to
solving a \emph{denoising} problem that does not depend on $\Hbf$ and imposes
piecewise-smoothess on the reconstruction~\cite{Cheng.Hofmann2011}.
%: 
%\begin{equation} \prox_{\tau \mathcal{R}_{\rm
%  TV}}(\zbf) \defn \argmin_{\xbf \in \R^N} \left\{\frac{1}{2}\|\xbf -
%  \zbf\|_{\ell_2}^2 + \tau \mathcal{R}_{\rm TV}(\xbf)\right\}\hspace{-0.23em}.
%\end{equation} 

%The gradient is represented with $2$ separate operators, $\Dbf \defn
%(\Dbf_x,\Dbf_y)$, computing finite-differences along each dimension of a 2D
%image. 

From a fundamental standpoint, the modular structure of FISTA and ADMM algorithms separates the prior model (specified by the proximal) from the
underlying physical model $\Hbf$. To develop more effective
regularizers than TV, researchers have thus modified the proximal operators
based on practical grounds (notably, the subsequent mean-squared-error (MSE)
performance) rather than analyticity. One class of algorithms called
``plug-and-play'' (PnP)~\cite{Venkatakrishnan.etal2013, Sreehari.etal2016,
Chan.etal2016, Kamilov.etal2017, Romano.etal2017} replaces the proximal step
with powerful denoising techniques such as BM3D~\cite{Dabov.etal2007}.  More
recently, motivated by the success of neural networks~\cite{LeCun.etal2015} in
image analysis applications~\cite{McCann.etal2017}, learning-based methods have
also been proposed for designing regularization strategies. One popular
approach is to \emph{unfold} a specific iterative reconstruction algorithm that
is derived for a TV-like regularization and consider a parametrized proximal
step instead of a fixed one.  Through the learning of parametrization
coefficients in a data-driven fashion, such algorithms have adapted the
regularizer to the underlying properties (deterministic and/or stochastic) of
the data~\cite{Barbu2009, Schmidt.Roth2014, Chen.etal2015,
Kamilov.Mansour2016, Mahapatra.etal2017}.

The efficiency of designing trainable regularizers is primarily determined by
the algorithm that is chosen to be unfolded. The major challenge is that many
proximal operators, such as that of TV, do not admit closed form solutions and
require additional iterative solvers for computation~\cite{Beck.Teboulle2009a,
Qin.etal2011}. This complication might limit the learning process to
differentiable models~\cite{Chen.etal2015}. Alternatively, ISTA-based schemes 
can be used without such confinements for learning proximals that are
simpler~\cite{Kamilov.Mansour2016}.  Using
variable-splitting~\cite{Afonso.etal2010}, ADMM-based learning has
addressed these proximal-related problems.  However, the final reconstruction
algorithm obtained by this formulation is efficient only for a restricted class
of forward models due to the inherent properties of
ADMM~\cite{Schmidt.Roth2014, Yang.etal2016}. Moreover, since variable-splitting
introduces auxiliary variables, such methods also require more memory, which
becomes a bottle-neck for large-scale imaging problems~\cite{Antipa.etal2018}. 

In this letter, we propose a new learning-based image reconstruction method
called the \emph{trainable} parallel proximal algorithm (TPPA). Our algorithm
extends the recently proposed fast parallel proximal algorithm
(FPPA)~\cite{Kamilov2017} to its data-adaptive variant. At its core, FPPA uses
a simple wavelet-domain soft-thresholding to compute the proximal of TV,
eliminating the need for an additional iterative solver. Building upon this
aspect, our framework: {\bf 1)} efficiently learns a TV-type regularization by
replacing the soft-thresholding function by a parametric representation that is
then learned for a given data-class, {\bf 2)} is general and does not put any
restrictions on the forward model $\Hbf$.  We also show that the training and
reconstruction processes share the same algorithmic structure, making TPPA's
end-to-end implementation very convenient. We apply the proposed
method to the problem of deconvolution in fluorescence microscopy. Our results
show that the learned regularization improves the deconvolution accuracy
compared to TV and PnP models.

%~\cite{Donoho1995, Mallat2009,
%Kamilov.etal2012}
 % Introduction
\section{Mathematical Background} \label{Sec:Method}

Our formalism starts with discussing the fundamentals of the FPPA method. This
is then followed by the derivation of our method, which is the data-driven
variant of FPPA.

%%%%%%%%%%%%%%%%%%%%%%%%%%%%%%%%%%%%%%%%%%%%%%%%%%%%%%%%%%%%%%%%%%%%%%%%%%%%%%%%
\subsection{FPPA for TV regularization} \label{Sec:FPPA}
%%%%%%%%%%%%%%%%%%%%%%%%%%%%%%%%%%%%%%%%%%%%%%%%%%%%%%%%%%%%%%%%%%%%%%%%%%%%%%%%
First, we provide some background on TV regularization via FPPA. The
method uses wavelets to define (and generalize) the TV regularizer.  To see
this, we first define a transform $\Wbf: \R^{N} \rightarrow \R^{N \times 4}$
that consists of the gradient operator $\Dbf = (\Dbf_x, \Dbf_y)$, as well as an
averaging operator ${\Abf = (\Abf_x, \Abf_y)}$. The averaging operator $\Abf$
computes pairwise averages of pixels along each dimension.  We rescale both
operators by $1/(2\sqrt{2})$ for notational convenience. Note that combining
these operators makes $\Wbf$ an invertible transform and it holds that
$\Wbf^\T\Wbf = \Ibf$, which is not the case for $\Dbf$ alone. However, note that
$\Wbf\Wbf^\T \neq \Ibf$ due to $\Wbf$ being redundant~\cite{Elad.etal2007}. 

$\Wbf$ can be rewritten as a union of four orthogonal transforms
$\{\Wbf_k\}_{k \in [1\dots4]}$, allowing $\Wbf$ to be
interpreted as the union of scaled and shifted Haar wavelets and scaling
functions~\cite{Mallat2009}. This viewpoint provides us with the central idea
of FPPA, which recasts the TV regularizer by using the four orthogonal Haar
transforms: 
\begin{equation} 
  \Rcal_{\rm TV}(\xbf) = \tau\sqrt{2} \sum_{k = 1}^4 \sum_{n \in \Hcal_k}
  \left|[\Wbf_k \xbf]_n\right|\text{.} 
\end{equation} 
$\Hcal_k \subset [1, \dots, N]$ is the set of all the detail
(\textit{i.e.}~difference) coefficients of the transform $\Wbf_k$.  This relationship
is then used to design the following updates at iteration $t$:
\begin{subequations} 
  \label{Eq:FPPA} 
  \begin{align} 
    &\sbf^t \leftarrow \mu_t\xbf^{t-1} + (1-\mu_t) \xbf^{t-2} \\ 
    &\zbf^t \leftarrow \sbf^t - \gamma_t \Hbf^\T(\Hbf\sbf^t-\ybf) \\ 
    &\xbf^t \leftarrow \Wbf^\T \Tcal(\Wbf\zbf^t, 2\sqrt{2}\tau\gamma_t), 
  \end{align} 
\end{subequations}
where the scalar soft-thresholding function 
\begin{equation} 
  \Tcal(z, \tau)
  \defn \sgn(z)\max(|z|-\tau, 0), 
\end{equation} 
is applied element-wise on the detail coefficients. As in the FISTA
implementation of TV (TV-FISTA)~\cite{Beck.Teboulle2009a}, the parameters
$\{\mu_t\}$ are set as~\cite{Nesterov1983}
\begin{equation}  
  \mu_t = 1 - \frac{1-q_{t-1}}{q_t}
  \text{, with } q_t = \frac{1}{2}(1+\sqrt{1+4q_{t-1}^2})
\end{equation} 
and $q_0 = 1$. Note that FPPA exploits the well-known connection between the
Haar wavelet-transform and TV, and it is closely related to a technique called
cycle spinning~\cite{Coifman.Donoho1995, Fletcher.etal2002, Steidl.etal2004,
Kamilov.etal2014a}.

The convergence rate of FPPA is given by~\cite{Kamilov2017} 
\begin{equation}
  \label{Eq:ConvRate}
  \Ccal(\xbf^t) - \Ccal(\xbf^\ast) \leq \frac{2}{\gamma(t+1)^2}
  \|\xbf^0-\xbf^\ast\|_{\ell_2}^2 + 4\gamma G^2, 
\end{equation}
where $\{\xbf^t\}$ are the iterates from~\eqref{Eq:FPPA}, $\Ccal$ is the true
TV cost functional, and $\xbf^\ast$ is a minimizer of $\Ccal$. This means that
for a constant step-size $\gamma > 0$, convergence can be established in the
neighborhood of the optimum, which can be made arbitrarily close by letting
$\gamma \rightarrow 0$. Additionally, the global convergence rate of FPPA
$O(1/t^2)$ matches that of TV-FISTA~\cite{Beck.Teboulle2009a}. 

FPPA that works with a fixed regularizer such as TV. The idea and convergence
of FPPA can be generalized to regularizers beyond TV by using other wavelet
transform and considering multiple resolutions.

%%%%%%%%%%%%%%%%%%%%%%%%%%%%%%%%%%%%%%%%%%%%%%%%%%%%%%%%%%%%%%%%%%%%%%%%%%%%%%%%
\section{Proposed Approach: Trainable Parallel Proximal Algorithm (TPPA)} 
\label{Sec:TPPA}
%%%%%%%%%%%%%%%%%%%%%%%%%%%%%%%%%%%%%%%%%%%%%%%%%%%%%%%%%%%%%%%%%%%%%%%%%%%%%%%%
We now present our method, which adapts the regularization to the data rather
than being designed for a fixed one. Given $\Hbf$ and $\Wbf$,  we see that the
shrinkage function solely determines the reconstruction. We have
noted that the TV reconstruction is strictly linked to the soft-thresholding
within the scheme outlined in~\eqref{Eq:FPPA}. However, the efficiency of a
shrinkage function varies with the type of object being
imaged~\cite{Bostan.etal2013}. This necessitates revisiting FPPA to obtain a
data-specific reconstruction algorithm.

Our model keeps $\{\Wbf_k\}_{k\in[1\dots4]}$ as the pairwise  
averages and differences and considers an iteration-dependent sequence of
shrinkage functions for each wavelet channel $k \in [1 \dots 4]$. We adopt
the following parametrization: 
\begin{equation} 
  \Tcal_k^t(x) =
  \sum_{p = -P}^P c_{kp}^t
  \varphi\left(\frac{x}{\Delta}-p\right)\hspace{-.23em}, 
\end{equation}
where $\{ c_{kp}^t \}$ are the expansion coefficients and
$\varphi$ is a basis function positioned on the grid ${\Delta[-P \dots P]
\subset \Delta \Z}$. We additionally reparametrize each step-size
$\gamma_t > 0$ with a scalar $\alpha_t \in \R$ and a one-to-one function 
\begin{equation} 
  \gamma = \phi(\alpha) = 
  \begin{dcases*}
    \e^{\alpha-1}  & if $\alpha \leq 1 \text{,}$\\ \alpha & \text{otherwise}.
  \end{dcases*} 
\end{equation} 
This representation facilitates automatic tuning of the step-sizes
$\{\gamma_t\}$ while ensuring their non-negativity. We note that the overall
parametrization can be restricted to its iteration-independent counterpart
(\emph{i.e.}~same set of parameters for each iteration). Moreover, by appropriately
constraining the parameters to lie in a well-characterized
subspace, the convergence rate given
in~\eqref{Eq:ConvRate} can be preserved. However, such constraints are
potentially restrictive on the reconstruction performance~\cite{Nguyen.etal2017}.

At iteration $t$, the TPPA updates are
\begin{subequations}
  \label{Eq:TPPA}
  \begin{align} 
    &\sbf^t \leftarrow \mu_t\xbf^{t-1} + (1-\mu_t) \xbf^{t-2} \\ 
    &\zbf^t \leftarrow \sbf^t - \phi(\alpha_t) \Hbf^\T(\Hbf\sbf^t-\ybf) \\ &
    \xbf^t \leftarrow \textstyle \sum_{k =1}^4 \Wbf_k^\T \Tcal_k^t(\Wbf_k\zbf^t), 
  \end{align}
\end{subequations}
where the scaling factors are absorbed into the coefficients. In contrast
to~\eqref{Eq:FPPA}, TPPA uses a sequence of adjustable shrinkage functions for
each $\Wbf_k$ in addition to self-tuning the step-size. More importantly,
compared to similar approaches based on
ADMM~\cite{Yang.etal2016,Nguyen.etal2017}, TPPA does not rely on $\Hbf^\T \Hbf$
being a structured matrix (such as block-circulant) for computational
efficiency.

\subsection{Training of Model Parameters}
We now consider determining our model parameters (\textit{i.e.}~shrinkage
functions and step-sizes) via an offline training.  Through a collection of
training pairs $\{(\xbf_\ell, \ybf_\ell)\}_{\ell \in [1 \dots L]}$, our goal is
to learn  
$$
\thetabm = \{\thetabm^t\}_{t \in [1 \dots T]},
$$
where $\thetabm^t \defn \{\alpha_t, \cbf^t\}$, with $\cbf^t =
\{c_{kp}^t\}_{k\in[1\dots4]\text{, }p\in[-P\dots P]}$ denoting the vector of
coefficients.  The total number of trainable parameters is 
$
{\rm dim}(\thetabm) = T+TK(2P+1).
$ 
We define the cost for parameter learning to be the mean squared error (MSE)
over the training data 
\begin{equation}
  \Ecal(\thetabm) = \frac{1}{2}\sum_{\ell = 1}^L \|\xbfhat(\thetabm, \ybf_\ell)
  - \xbf_\ell\|_{\ell_2}^2,
\end{equation} 
where $\xbfhat(\thetabm, \ybf)$ is the output of~\eqref{Eq:TPPA} for a given
measurement vector $\ybf_{\ell}$ and set of parameters $\thetabm$ after a fixed
number of iterations $T$. The learned parameters are thus obtained via 
\begin{equation}
  \thetabm^{\star} = \underset{\thetabm}{\arg \min} \,\,\, \Ecal(\thetabm)\text{.}
  \label{Eq:OptimizationTheta}
\end{equation}
The implication of~\eqref{Eq:OptimizationTheta} is immediate: Given
a fixed computation cost (that is expectedly cheaper than that for TV), the
shrinkages are optimized to maximize the reconstruction accuracy over
the training dataset. 

Note that the optimization problem in~\eqref{Eq:OptimizationTheta} is smooth
and hence first-order optimization methods are convenient. We use the gradient
descent algorithm with Nesterov's acceleration scheme~\cite{Nesterov1983} (see
Algorithm~\ref{Alg:Training}\footnote{Note that iterates of the parameters
are represented by $\thetabm^{(i)}$ to distinguish from $\thetabm^{t}$. }).

We now explain how the gradient of the cost function
in~\eqref{Eq:OptimizationTheta} is derived. We denote the gradient by $\nabla
\Ecal$ and rely on backpropagation~\cite{Demuth.etal2014} for obtaining its
analytical expression. Here, we point out the main aspects of our derivation since
such calculations are lengthy. First, we define the residual term $
\rbf^t \defn \left[ \partial \Ecal / \partial \xbf^t  \right]^\T
$
and use the chain rule to get: 
\begin{align}
  \rbf^{t-2} &= 
\left[ \frac{\partial \Ecal}{\partial \xbf^{t-2}} \right]^\T \nonumber \\
&= 
\left[ \frac{\partial \xbf^{t-1}}{\partial \xbf^{t-2}}\right]^\T \rbf^{t-1} + 
\left[ \frac{\partial \xbf^{t}}{\partial \xbf^{t-2}}\right]^\T \rbf^{t}\text{.}
  \label{Eq:ResidualRecursion}
\end{align} Using matrix calculus, the derivatives as thus
$$
\frac{\partial \xbf^{t-1}}{\partial \xbf^{t-2}} = 
\mu_{t-1} \Wbf^\T {\rm diag}\left(\Tcal_{t-1}^{\prime}(\ubf^t) \right) 
\Wbf \left( \Ibf - \gamma_{t-1} \Hbf^{\T}\Hbf \right)\text{,} 
$$
where $\ubf^t = \Wbf \zbf^t$. Similarly, we compute
$$
\frac{\partial \xbf^{t}}{\partial \xbf^{t-2}} = 
(1 - \mu_{t})\Wbf^\T {\rm diag}\left(\Tcal_{t}^{\prime}(\ubf^t) \right) 
\Wbf \left( \Ibf - \gamma_{t} \Hbf^{\T}\Hbf \right)\text{.} 
$$ 
As for the derivatives of the training parameters, we use the chain rule once
again to attain the following:
\begin{align*}
\left[ \frac{\partial \Ecal}{\partial \alpha^t} \right]^\T &= 
-\phi^{\prime}(\alpha^t) (\Hbf \sbf^t -\ybf) \Hbf \Wbf^\T 
{\rm diag}\left(\Tcal_t^{\prime}(\ubf^t) \right) \Wbf \rbf^t
\text{;} \\
\left[ \frac{\partial \Ecal}{\partial \cbf_k^t} \right]^\T &= 
\left( \Phibf_k^t \right)^\T\Wbf_k\rbf^t\text{,}
\end{align*}
where $\Phibf_k^t \in \R^{N \times (2P+1)}$ is the matrix representation of the
basis functions in the sense that $\Tcal_k^t(\ubf_k^t) = \Phibf_k^t \cbf_k^t$.
Considering these partial derivatives along
with~\eqref{Eq:ResidualRecursion}, we obtain the scheme described in
Algorithm~\ref{Alg:BackProp} to compute the backpropagation. Finally, we note
that the reconstruction in~\eqref{Eq:TPPA} and Algorithms~\ref{Alg:Training}
and~\ref{Alg:BackProp} essentially share the same structure
(\textit{i.e.}~gradient descent with acceleration). This makes the proposed model
convenient, since the computational implementation can be reused. 

\begin{algorithm}[t!]
	\caption{Parameter training}
	\label{Alg:Training}
	{\bf Input:} a training pair $(\xbf_{\ell},\ybf_{\ell})$, learning rate
	$\nu$, number of training iterations $I $.\\
	{\bf Output:} optimized parameters $\thetabm^{\star}$. 
	\\
	Initialize: $\thetabm^{(0)}$ and $\phibm^{(0)}$  \\
	Set: $q_0 =1$. \\
	For $i = 1, 2, \dots, I$, compute  
	\begin{itemize}
	  \item[] $\thetabm^{(i)} \leftarrow \phibm^{(i-1)} - \nu \nabla \Ecal(\phibm^{(i-1)} )$ (use Algorithm~\ref{Alg:BackProp}),	
	  \item[] $q_i \leftarrow \left( 1+ \sqrt{1+4q_{t-1}^2}\right)/2$,
	  \item[] $\phibm^{(i)} \leftarrow \thetabm^{(i)}  + (q_{i-1}-1/q_i)(\thetabm^{(i)} -\thetabm^{(i-1)})$,
	\end{itemize}
	and return $\thetabm^{(I)}$.
\end{algorithm}

\begin{algorithm}[t!]
  \caption{Backpropagation for Algorithm~\ref{Alg:Training}}
	\label{Alg:BackProp}
	{\bf Input:} a training pair $(\xbf_{\ell},\ybf_{\ell})$, the set
	of parameters $\thetabm$, number of TPPA iterations $T$.\\
	{\bf Output:} components of the gradient $\nabla \Ecal$. 
	\\
	Set: $\mu^{T+1} = 1$, $\vbf^{T+1} = \mathbf{0}$, and 
	    $\rbf^T = (\hat{\xbf}(\thetabm,\ybf_{\ell}) - \xbf_{\ell}).$ \\
	For $t = T, T-1, \dots, 1$, compute  
	\begin{itemize}
	  \item[] $\bbf^t \leftarrow \Wbf^\T {\rm
	  diag}\left(\Tcal_{t}^{\prime}(\ubf^t) \right) \Wbf \rbf^t$,

	\item[] $\vbf^t \leftarrow \bbf^t - \gamma_t \Hbf^\T \Hbf \bbf^t$,
	\item[] $\rbf^{t-1} \leftarrow \mu_t \vbf^t + (1- \mu_{t+1})\vbf^{t+1}$,
\end{itemize}
and store 
\begin{itemize}

	\item[] $\left[ \frac{\partial \Ecal}{\partial \alpha^t}\right]^\T = 
	   -\phi^{\prime}(\alpha^t) (\Hbf \sbf^t - \ybf_{\ell})^\T \Hbf \bbf^t$,
	 \item[] $\left[ \frac{\partial \Ecal}{\partial \cbf_k^t}\right]^\T = 
	   \left( \Phibf_k^t \right)^\T \Wbf_k \rbf^t$ ($k=1,\dots,4$).
	\end{itemize}
\end{algorithm}
 % Proposed Method
%%%%%%%%%%%%%%%%%%%%%%%%%%%%%%%%%%%%%%%%%%%%%
%% Experimental results
%%%%%%%%%%%%%%%%%%%%%%%%%%%%%%%%%%%%%%%%%%%%%

\section{Numerical Results}
\label{Sec:NumericalResults}

%%%%%%%%%%%%%%%%%%%%%%%%%%%%%%%%%
\begin{figure*}[t]
  \centering
  \includegraphics[scale=0.5]{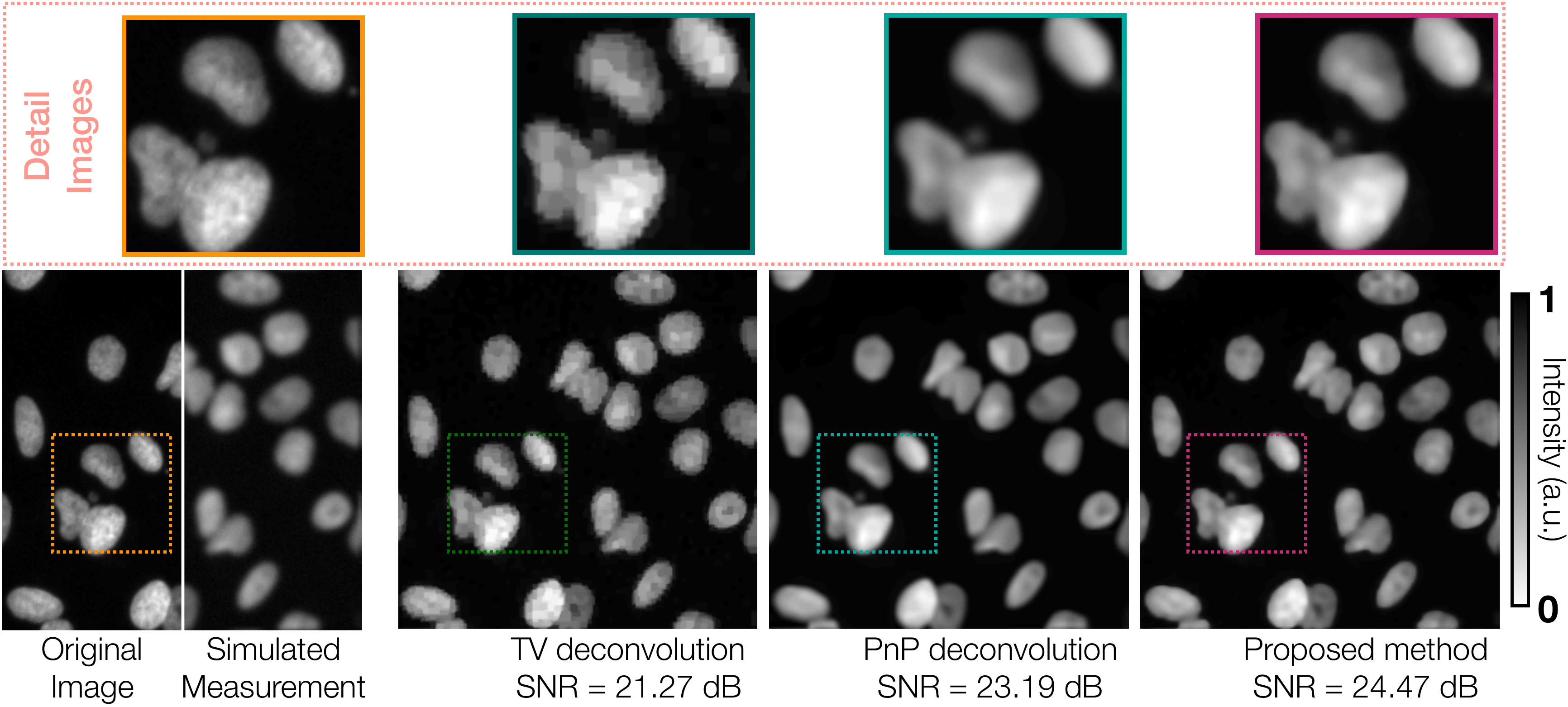}
  \caption{An example of 2D deconvolution with fluorescence microscopy images,
    given a known 9$\times$9 Gaussian PSF: The ground-truth image (left)
    illustrates the nucleus of a group of U2OS cells. Our learning-based
    reconstruction preserves homogeneity of the background and more of the
    texture, increasing the SNR. See text for further details.}
  \label{fig:VisualResults}
\end{figure*}
%%%%%%%%%%%%%%%%%%%%%%%%%%%%%%%%%

%%%%%%%%%%%%%%%%%%%%%%%%%%%%%%%%%
%\begin{figure}[h]
%  \centering
%  \includegraphics[scale=0.36]{training_performance.pdf}
%  \caption{Progress of the training: The error as a function of the iterations has decreased and
%  the deconvolution performance has stabilized. See text for further details.}
%  \label{fig:TrainingPerformance}
%\end{figure}
%%%%%%%%%%%%%%%%%%%%%%%%%%%%%%%%%

We now present \textit{in silico} experiments corroborating TPPA, with
deconvolution of fluorescence microscopy images where the point spread function
(PSF) of the microscope is approximated by a Gaussian kernel of variance 2
pixels. The imaging process is assumed not to be photon-limited; noise is
modeled as additive white Gaussian noise (AWGN) of 30 dB SNR. 

%Our implementation assumes periodic boundary conditions.  

Fluorescence microscopy images of human bone osteosarcoma epithelial cells
(U2OS Line)\footnote{The dataset consists of multi-color images where each
  color channel depicts different flurophores targeted to different organelles
  or cellular components.  In our simulations, we use the channel corresponding
to the blue color which targets the cell nucleus.}, from ~\cite{Bray.etal2017}
are used as our ground-truth data. All images' intensity are scaled between 0
and 1. To generate the training pairs, we use 100 images and apply the forward
model to a single patch (per image) of size 64$\times$64 extracted around the
center of the field-of-view.  Once the images chosen for training are excluded,
we select a different 20 images of size 256$\times$256 as a validation set. 

\begin{table}
\caption{Average deconvolution performance  (on the validation set) of the
methods considered in the experiments. Numbers indicate SNR in decibel units. }
\begin{center}
     \begin{tabular*}{0.475\textwidth}[t!]{ c  c  c  c}
    \toprule
                  &  \multicolumn{3}{c}{Deconvolution Algorithm}  \\
		     \cmidrule(r){2-4}
    PSF Kernel Size & TV     & PnP (using BM3D) & Proposed Method \\ \hline
    5$\times$5      & 21.99  & 24.69            & \bf{24.96}      \\ 
    9$\times$9      & 20.77  & 22.33            & \bf{22.57}      \\ 
    
    \bottomrule
  \end{tabular*}
  \end{center}
  \label{tab:DeconResults}  
\end{table}

Learning is carried out by using Algorithm~\ref{Alg:Training} with 200 iterations
(that is $I=200$) with $\nu = 5 \times × 10^{-4}$. We set the
number of layers for TPPA as $T=10$. The shrinkage functions are
parametrized by $10^{3}$ equally-spaced cubic B-splines over the dynamic range
of $\Wbf\xbf$. All shrinkages are initialized with the identity operator.
Finally, we note that $\alpha_0 = 1/\| \Hbf^{\T}\Hbf\|_2^2$ and $\xbf^0 =
\mathbf{0}\in\mathbb{R}^N$.

As a baseline comparison, we consider TV regularization implemented using FPPA
described in~\eqref{Eq:FPPA}. The algorithm is run until either 100 iterations
is reached or $\| \xbf_t - \xbf_{t-1} \|_2 / \| \xbf_{t-1} \|_2 \leq 10^{-6}$
is satisfied. We also compare against the PnP model where the proximal of TV is
replaced by BM3D~\cite{Dabov.etal2007}. The latter is implemented using 10
FISTA iterations (same as the number of layers in TPPA) and all methods use a zero
initialization.  For each validation image, we optimize the
regularization parameters for both algorithms (by using an oracle) for the
best-possible SNR performance.  Average SNRs of the reconstruction are reported
in Table~\ref{tab:DeconResults} for different sizes of the blur kernel. 

The results show that the accuracy of our model is better than the other
algorithms considered. In particular, the SNR performance provided
by TPPA is significantly better than that of TV. Furthermore, visual
inspection of the reconstructions reveals that the TV deconvolution creates the
characteristic blocky artifacts at textured regions (see
Figure~\ref{fig:VisualResults}). Since the successive sequence of shrinkage
functions are adapted to the underlying features of the training data, one
notices that these artifacts are reduced for TPPA. Our
method also renders the boundary of the nucleus more faithfully and provides a
homogeneous background. These observations confirm the efficiency of our
data-specific approach (in terms of deconvolution quality) and highlight its
potential importance in practical scenarios. 
 % Experimental Results
%%%%%%%%%%%%%%%%%%%%%%%%%%%%%%%%%%%%%%%%%%%%%
%% Conclusion
%%%%%%%%%%%%%%%%%%%%%%%%%%%%%%%%%%%%%%%%%%%%%

\section{Conclusion}
\label{Sec:Conclusion}

We developed a learning-based algorithm for
linear inverse problems that is in the spirit of TV regularization. Our
approach, TPPA, has enabled us to move away from the soft-thresholding operator
(with a fixed threshold value at all iterations) to a collection of
parametrized shrinkage functions that are optimized (in MSE sense) for a
training set.  Compared to TV regularization and PnP technique, our
deconvolution simulations demonstrate that advantages of TPPA in terms of
accuracy.

%built upon the existing architecture of FPPA where the
%proximal operator corresponding to TV is computed in a non-iterative fashion.
%This has
 % Conclusion

%%%%%%%%%%%%%%%%%%%%%%%%%%%%%%%%%%%%%%%%%%%%%
%% Bibliography
%%%%%%%%%%%%%%%%%%%%%%%%%%%%%%%%%%%%%%%%%%%%%

\bibliographystyle{IEEEtran}

% Generated by IEEEtran.bst, version: 1.14 (2015/08/26)

\end{document}